\documentclass[11pt]{article}
\pdfoutput=1
\usepackage{naaclhlt2015}
\usepackage{amsmath}
\usepackage{natbib}
\usepackage{multirow}
\usepackage{multicol}
\usepackage{mathtools}
\usepackage{fixltx2e}
\usepackage{tipa}
\usepackage{tikz}
\usetikzlibrary{arrows,calc,chains,fit,matrix,patterns,positioning,shapes,snakes}
\usepackage{pgfplots}
\usepackage{url}
\usepackage{xspace}
\usepackage{latexsym}
\usepackage{mathptmx}
\usepackage[scaled=.90]{helvet}
\usepackage{courier}

\newcommand{\acronym}[1]{\textsc{#1}\xspace}

\newcommand{\POS}{\acronym{pos}}
\newcommand{\Ngram}{\emph{N}-gram\xspace}
\newcommand{\ngram}{\emph{n}-gram\xspace}
\newcommand{\ngrams}{\emph{n}-grams\xspace}
\newcommand{\bfngram}{\textbf{\emph{n}}-gram\xspace}

\newcommand{\WSJ}{\acronym{wsj}}

\newcommand{\syntacticngrams}{Syntactic Ngrams\xspace}
\newcommand{\sancl}{\acronym{sancl}}
\newcommand{\pennconverter}{pennconverter\xspace}
\newcommand{\UAS}{\acronym{uas}}

\newcommand{\LTH}{\acronym{lth}}

\newcommand{\MXPOST}{\acronym{mxpost}}
\newcommand{\MST}{\acronym{mst}}
\newcommand{\BLLIP}{\acronym{bllip}}
\newcommand{\BASE}{\acronym{base}}
\newcommand{\WEBT}{\acronym{web}}
\newcommand{\BOOKS}{\acronym{bks}}
\newcommand{\SYNTACTIC}{\acronym{syn}}
\newcommand{\COMBINED}{\acronym{comb}}
\newcommand{\ANSWERS}{\acronym{ans}}
\newcommand{\NEWSGROUPS}{\acronym{ngs}}
\newcommand{\REVIEWS}{\acronym{rev}}
\newcommand{\EWT}{\acronym{ewt}}
\newcommand{\AVG}{\acronym{avg}}
\newcommand{\NP}{\acronym{np}}
\newcommand{\PP}{\acronym{pp}}

\newcommand{\sent}[1]{{\textit{\textsf{#1}}}}

\newcommand{\schwalab}{\textschwa-lab\xspace}

\newcommand{\OCR}{\acronym{ocr}}
\newcommand{\mstparser}{MSTParser\xspace}

\DeclarePairedDelimiter\floor{\lfloor}{\rfloor}

\title{Web-scale Surface and Syntactic \bfngram Features for Dependency Parsing}

\author{
  Dominick Ng$^{\diamond}$ \hspace{3mm} Mohit Bansal$^{\dagger}$\hspace{3mm}
  James R. Curran$^{\diamond}$ \\\\
  \begin{tabular}[c]{ccc}
    \rm $^{\diamond}$\schwalab, School of Information Technologies & \rm $^{\dagger}$ Toyota Technological Institute at Chicago \\
    \rm University of Sydney & \rm Chicago, IL 60637 \\
    \rm NSW 2006, Australia & \rm \\
    \rm \small {\tt \{dominick.ng,james.r.curran\}@sydney.edu.au} & \small {\tt mbansal@ttic.edu} \\
  \end{tabular}
}

\date{}

\begin{document}

\maketitle

\begin{abstract}
  We develop novel first- and second-order features for dependency parsing
  based on the Google \syntacticngrams corpus, a collection of subtree counts
  of parsed sentences from scanned books. We also extend previous work
  on surface \ngram features from Web1T to the Google Books corpus and from
  first-order to second-order, comparing and analysing performance over
  newswire and web treebanks.

  Surface and syntactic \ngrams both produce substantial and complementary
  gains in parsing accuracy across domains. Our best system combines the two
  feature sets, achieving up to 0.8\% absolute \UAS improvements on newswire
  and 1.4\% on web text.
\end{abstract}

\section{\label{sect:intro}Introduction}

Current state-of-the-art parsers score over 90\% on the standard newswire
evaluation, but the remaining errors are difficult to overcome using only the
training corpus. Features from \ngram counts over resources like Web1T
\citep{brants:06a} have proven to be useful proxies for syntax
\citep{bansal:11a,pitler:12b}, but they enforce linear word order, and are
unable to distinguish between syntactic and non-syntactic co-occurrences.
Longer \ngrams are also noisier and sparser, limiting the range of potential
features.

% Both unlabeled and labeled datasets have been identified as potential
% sources of syntactic cues to address the remaining parser errors.
% \citet{bansal:11a}
% long range for ngrams?
% investigate any possible spurious features for mohit

In this paper we develop new features for the graph-based \mstparser
\citep{mcdonald:06a} from the Google \syntacticngrams corpus
\citep{goldberg:13a}, a collection of Stanford dependency subtree counts. These
features capture information collated across millions of subtrees produced by
a shift-reduce parser, trading off potential systemic parser errors for data
that is better aligned with the parsing task. We compare the performance of
our syntactic \ngram features against the surface \ngram features of
\citet{bansal:11a} in-domain on newswire and out-of-domain on the English Web
Treebank \citep{petrov:12b} across CoNLL-style (\LTH) dependencies. We also
extend the first-order surface \ngram features to second-order, and compare
the utility of Web1T and the Google Books Ngram corpus \citep{lin:12a} as
surface \ngram sources.

We find that surface and syntactic \ngrams provide statistically significant
and complementary accuracy improvements in- and out-of-domain. Our best \LTH
system combines the two feature sets to achieve 92.5\% unlabeled attachment
score (\UAS) on newswire and 85.2\% \UAS averaged over web text on a baseline
of 91.7\% and 83.8\%. Our analysis shows that the combined system is able to
draw upon the strengths of both surface and syntactic features whilst avoiding
their weaknesses.

\section{\label{sect:syntactic-ngrams}Syntactic \bfngram Features}

% \begin{figure*}[ht]
% \begin{center}
%   \begin{verbatim}hold   hold/VBP/ROOT/0 a/DT/det/3 hearing/NN/dobj/1   174   1920,3...\end{verbatim}
% \end{center}
% \caption{\label{figure:syntactic-ngrams-corpus-line}A truncated line from the
%   extended arcs set of the Google \syntacticngrams corpus.
%   The fields are head word, the syntactic \ngram (word/\POS/label/head),
%   total frequency, and frequency by year(s).}
% \end{figure*}

The Google \syntacticngrams English (2013) corpus\footnote{\tiny
\url{commondatastorage.googleapis.com/books/syntactic-ngrams}} contains
counts of dependency tree fragments over a 345 billion word selection of the
Google Books data, parsed with a beam-search shift-reduce parser and Stanford
dependencies \citep{goldberg:13a}. The parser is trained over substantially
more annotated data than typically used in dependency parsing.

Unlike surface \ngrams, syntactic \ngrams are not restricted to linear word
order or affected by non-syntactic co-occurrence. Given a head-argument
ambiguity, we extract different combinations of word, \POS tag, and
directionality, and search the \syntacticngrams corpus for matching subtrees.
To reduce the impact of this search during run time, we extract all possible
combinations in the training and test corpora ahead of time and total the
frequencies of each configuration, storing these in a lookup table that is
used by the parser at run-time to compute feature values. We did not use any
features based on the dependency label as these are assigned in a separate
pass in \mstparser.

\begin{table}[t]
\begin{small}
\begin{center}
  \begin{tabular}{l|r|r}
    \hline
    Feature Lookup & Count & \acronym{Bucket} \\
    \hline\hline
    \sent{hold} (head) & 80,129k & 4 \\
    \sent{hearing} (arg) & 7,839k & 4 \\
    \sent{hold} $\rightarrow$ \sent{hearing}  & 15k & 3 \\
    \sent{hold} $\rightarrow$ \sent{hearing} (head left) & 15k & 3 \\
    \hline
    \texttt{VB} (head) & 20,996,911k & 5 \\
    \texttt{NN} (arg) & 22,163,825k & 5 \\
    \texttt{VB} (child right) & 6,261,484k & 5 \\
    \texttt{NN} (head left) & 15,478,472k & 5 \\
    \texttt{VB} $\rightarrow$ \texttt{NN} & 1,784,891k & 5 \\
    \texttt{VB} $\rightarrow$ \texttt{NN} (head left) & 1,437,932k & 5 \\
    \hline
    \sent{hold} $\rightarrow$ \texttt{NN} & 7,362k & 4 \\
    \sent{hold} $\rightarrow$ \texttt{NN} (head left) & 6,248k & 4 \\
    \texttt{VB} $\rightarrow$ \sent{hearing} & 396k & 3 \\
    \texttt{VB} $\rightarrow$ \sent{hearing} (head left) & 354k & 3 \\
    \hline
  \end{tabular}
  \caption{\label{table:syntactic-ngram-counts} Syntactic \ngram features,
  their counts in the extended arcs dataset, and the bucketed count
  for the \emph{hold} and \emph{hearing} dependency.}
\end{center}
\end{small}
\end{table}

\begin{figure}
\begin{center}
  \begin{tikzpicture}[font=\sffamily\normalsize,
    start chain, node distance=1mm,
    every node/.style={text height=1.5ex, text depth=.25ex,
    minimum height=2em, inner sep=2.8, text centered, on chain}]
    \tikzstyle{deplabelmid}=[black,thick,minimum height=0,text height=0.5ex, text depth=0ex,inner sep=1, outer sep=1,fill=white,font=\scriptsize]
    \tikzstyle{deplabel}=[black,thick,minimum height=0,text height=0.5ex, text depth=0ex,inner sep=1, outer sep=1,auto,font=\scriptsize]
    \tikzstyle{deplabelrev}=[black,thick,minimum height=0,text height=0.5ex, text depth=0ex,inner sep=1, outer sep=1,auto,swap,font=\scriptsize]
    \node (0) {\textit{could}};
    \node (1) {\textbf{hold}};
    \node (2) {\textit{a}};
    \node (3) {\textit{public}};
    \node (4) {\textbf{hearing}};
    \node (5) {next};
    \node (6) {\textit{week}};
    \draw[->,out=160,in=20,blue,dashed] (1.north)++(-0.15,0) to node[deplabel,swap,pos=0.5]{aux} (0.north);
    \draw[->,out=48,in=125,very thick,red] (1.north)++(0.25,0) to
    node[deplabel,pos=0.5]{dobj} (4.north);
    \draw[->,out=45,in=135,blue,dashed] (1.north)++(0.1,0) to node[deplabelmid,pos=0.45]{tmod} (6.north);
    \draw[->,out=145,in=35,blue,dashed] (4.north)++(-0.15,0) to node[deplabel,pos=0.56,swap]{det} (2.north);
    \draw[->,out=165,in=15,blue,dashed] (4.north)++(-0.3,0) to node[deplabel,pos=0.6,swap]{amod} (3.north);
  \end{tikzpicture}
  \caption{\label{figure:context} The paraphrase-style context words around
  \sent{hold}$\rightarrow$\sent{hearing} in a syntactic \ngram. Context words
  are italicized and their arcs dashed.}
\end{center}
\end{figure}
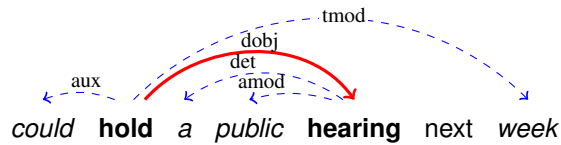

Table~\ref{table:syntactic-ngram-counts} summarizes the first-order features
extracted from the dependency \sent{hold} $\rightarrow$ \sent{hearing}
depicted in Figure~\ref{figure:context}. The final feature encodes the \POS
tags of the head and argument, directionality, the binned distance between the
head and argument, and a bucketed frequency
of the syntactic \ngram calculated as per Equation~\ref{equation:bucket},
creating bucket labels from 0 in increments of 5 (0, 5, 10, etc.).
\begin{align} \text{bucket} & = \floor*{\frac{\log_2(\sum \text{frequency})}{5}} \times 5
  \label{equation:bucket}
\end{align}
Additional features for each bucket value up to the maximum are also encoded.
We also develop paraphrase-style features like those of \citet{bansal:11a}
based on the most frequently occurring words and \POS tags before, in between,
and after each head-argument ambiguity (see
Section~\ref{sect:first-order-surface-ngrams}). Figure~\ref{figure:context}
depicts the potential context words available the \sent{hold} $\rightarrow$
\sent{hearing} dependency.

We experiment with a number of second-order features, mirroring those
extracted for surface \ngrams in
Section~\ref{sect:second-order-surface-ngrams}. We extract all triple and
sibling word and \POS structures considered by the parser in the training and
test corpora (following the factorization depicted in
Figure~\ref{figure:second-order}), and counted their frequency in the
\syntacticngrams corpus. Importantly, we require that matching subtrees in the
\syntacticngrams corpus maintain the position of the parent relative to its
children. We generate separate features encoding the word and \POS tag
variants of each triple and sibling structure. 
% \subsection{Working with syntactic \bfngrams}

Similar to the surface \ngram features (Section~\ref{sect:surface-ngrams}),
counts for our syntactic \ngram features are pre-computed to improve the
run-time efficiency of the parser. Experiments on the development set led to a
minimum cutoff frequency of 10,000 for each feature to avoid noise
from parser and \OCR errors.
% The \syntacticngrams corpus requires more
% intricate processing due to the format of the data.
% Figure~\ref{figure:syntactic-ngrams-corpus-line} gives an example line taken
% from the extended arcs dataset. The fields are the head word of the syntactic
% \ngram, the \ngram itself, the total frequency of the \ngram, and a frequency
% breakdown by year. The \ngram format is word/\POS/label/head with a 1-based
% indexing scheme; a 0 head index indicates the root of the subtree. The words
% in the \ngram maintain their original sentence ordering, but intervening words
% without a dependency arc to any word in the \ngram are omitted
% \citep{goldberg:13a}.
% Each subtree in the Google \syntacticngrams corpus must appear at least 10
% times, as opposed to the higher cutoff of 40 for both of the surface \ngram
% corpora in this work. We experimented with a frequency cutoff for our
% syntactic \ngram features to combat the low cutoff as well as potential noise
% from parser and \OCR errors. Tuning over the \WSJ development set led us to
% set a minimum cutoff frequency of 10,000 for each feature: the summed
% frequency count for each extracted subtree in the \syntacticngrams corpus had
% to be at least 10,000 for the feature to be used in the parser.

\section{\label{sect:surface-ngrams}Surface \bfngram Features}

\citet{bansal:11a} demonstrate that features generated from bucketing simple
surface \ngram counts and collecting the top paraphrase-based contextual words
over Web1T are useful for almost all attachment decisions, boosting dependency
parsing accuracy by up to 0.6\%. However, this technique is restricted to
counts based purely on the linear order of the adjacent words, and is unable
to incorporate disambiguating information such as \POS tags to avoid spurious
counts. \citet{bansal:11a} also tested only on in-domain text, though these
external count features should be useful out of domain.

We extract \citet{bansal:11a}'s affinity and paraphrase-style first-order
features from the Google Books English Ngrams corpus, and compare their
performance against Web1T counts. Both corpora are very large, contain
different types of noise, and are sourced from very different underlying
texts. We also extend \citeauthor{bansal:11a}'s affinity and paraphrase
features to second-order.

\subsection{Surface \bfngram Corpora}

The Web1T corpus contains counts of 1 to 5-grams over 1 trillion words of web
text \citep{brants:06a}. Unigrams must appear at least 200 times in the source
text before being included in the corpus, while longer \ngrams have a cutoff
of 40. \citet{pitler:10b} has documented a number of sources of noise in the
corpus, including duplicate sentences (such as legal disclaimers and
boilerplate text), disproportionately short or long sentences, and primarily
alphanumeric sentences.

The Google Books Ngrams English corpus (2012) contains counts of 1 to 5-grams
over 468 billion words sourced from scanned books published across three
centuries \citep{michel:11a}. A uniform cutoff of 40 applies to all \ngrams in
this corpus. This corpus is affected by the accuracy of \OCR and digitization
tools; the changing typography of books across time is one issue that may
create spurious co-occurrences and counts \citep{lin:12a}.

\subsection{\label{sect:first-order-surface-ngrams}First-order surface \bfngram features}

Affinity features rely on the intuition that frequently co-occurring words in
large unlabeled text collections are likely to be in a syntactic relationship
\citep{nakov:05a,bansal:11a}. \Ngram resources such as Web1T and Google Books
provide large offline collections from which these co-occurrence statistics
can be harvested; given each head and argument ambiguity in a training and
test corpus, the corpora can be linearly scanned ahead of parsing time to
reduce the impact of querying in the parser. When scanning, the head and
argument word may appear immediately adjacent to one another in linear order
(\acronym{contig}), or with up to three intervening words (\acronym{gap1},
\acronym{gap2}, and \acronym{gap3}) as the maximum \ngram length is five. The
total count is then discretized as per Equation~\ref{equation:bucket}
previously.

The final parser features include the \POS tags of the potential head and
argument, the discretized count, directionality, and the binned length of the
dependency. Additional cumulative features are generated using each
bucket from the pre-calculated up to the maximum bucket size.

Paraphrase-style surface \ngram features attempt to infer attachments
indirectly. \citet{nakov:05a} propose several static patterns to resolve a
variety of nominal and prepositional attachment ambiguities. For
example, they give the example of sentence (1) below, paraphrase it into
sentence (2), and examine how frequent the paraphrase is. If it should happen
sufficiently often, this serves as evidence for the nominal attachment to
\sent{demands} in sentence (1) rather than the verbal attachment to
\sent{meet}.

\begin{enumerate}
  \setlength{\itemsep}{1pt}
  \setlength{\parskip}{1pt}
  \item \sent{meet demands from customers}
  \item \sent{meet the customers demands}
\end{enumerate}

In \citet{bansal:11a}, paraphrase features are generated for all
full-parse attachment ambiguities from the surface \ngram corpus. For each
attachment ambiguity, 3-grams of the form ($\star$ $q_1$ $q_2$), ($q_1$
$\star$ $q_2$), and ($q_1$ $q_2$ $\star$) are extracted, where $q_1$ and $q_2$
are the head and argument in their linear order of appearance in the original
sentence, and $\star$ is any single context word appearing before, in between,
or after the query words. Then the most frequent words appearing in each of
these configurations for each head-argument ambiguity is encoded as a feature
with the \POS tags of the head and argument\footnote{The top 20 words in
between and top 5 words before and after are used for all paraphrase-style
features in this paper.}.

Given the arc \sent{hold} $\rightarrow$ \sent{hearing} in
Figure~\ref{figure:second-order}, \sent{public} is the most frequent word
appearing in the \ngram (\sent{hold} $\star$ \sent{hearing}) in Web1T. Thus,
the final encoded feature is \POS(hold) $\wedge$ \POS(hearing) $\wedge$
\sent{public} $\wedge$ \texttt{mid}. Further generalization is achieved by
using a unigram \POS tagger trained on the \WSJ data to tag each context word,
and encoding features using each unique tag of the most frequent
context words.

\subsection{\label{sect:second-order-surface-ngrams}Second-order surface \bfngram features}

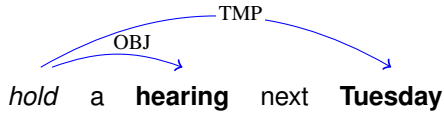
\begin{figure}[t]
\begin{center}
  \begin{tikzpicture}[font=\sffamily\normalsize,
    start chain, node distance=2mm,
    every node/.style={text height=1.5ex, text depth=.25ex,
    minimum height=2em, inner sep=2.8, text centered, on chain}]
    \tikzstyle{deplabelmid}=[black,thick,minimum height=0,text height=0.5ex, text depth=0ex,inner sep=1, outer sep=1,fill=white,font=\scriptsize]
    \tikzstyle{deplabel}=[black,thick,minimum height=0,text height=0.5ex, text depth=0ex,inner sep=1, outer sep=1,auto,font=\scriptsize]
    \tikzstyle{deplabelrev}=[black,thick,minimum height=0,text height=0.5ex, text depth=0ex,inner sep=1, outer sep=1,auto,swap,font=\scriptsize]
    \node (7) {\textit{hold}};
    \node (8) {a};
    \node (9) {\textbf{hearing}};
    \node (10) {next};
    \node (11) {\textbf{Tuesday}};
    \draw[->,out=20,in=160,blue] (7.north)++(0.25,0) to node[deplabel,pos=0.5]{OBJ} (9.north);
    \draw[->,out=30,in=150,blue] (7.north)++(0.1,0) to node[deplabelmid,pos=0.45]{TMP} (11.north);
    % \draw[->,out=160,in=20,gray,dashed] (9.north)++(-0.2,0) to node[deplabel,pos=0.56,swap,gray]{NMOD} (8.north);
    % \draw[->,out=160,in=20,gray,dashed] (11.north)++(-0.2,0) to node[deplabel,pos=0.56,swap,gray]{NMOD} (10.north);
  \end{tikzpicture}
  \caption{\label{figure:second-order} The second-order factorization used in
  \mstparser, with a parent and two adjacent children.}
\end{center}
\end{figure}

We extend the first-order surface \ngram features to new features over
second-order structures. We experimented with triple and sibling features,
reflecting the second-order factorization used in \mstparser (see
Figure~\ref{figure:second-order}).

As with first-order features, we convert all triple and sibling structures
from the training and test data into query \ngrams, maintaining their linear
order. In Figure~\ref{figure:second-order}, the corresponding \ngrams are
\sent{hold hearing Tuesday}, and \sent{hearing Tuesday}. We then scan the
\ngram corpus for each query \ngram and sum the frequency of each.
Frequencies are summed over each configuration (including intervening words)
that the query \ngram may appear in, as depicted below.
\begin{multicols}{2}
\begin{itemize}
  \setlength{\itemsep}{1pt}
  \setlength{\parskip}{1pt}
  \item ($q_1$ $q_2$ $q_3$)
  \item ($q_1$ $\star$ $q_2$ $q_3$)
  \item ($q_1$ $q_2$ $\star$ $q_3$)
  \item ($q_1$ $\star$ $q_2$ $\star$ $q_3$)
  \item ($q_1$ $\star$ $\star$ $q_2$ $q_3$)
  \item ($q_1$ $q_2$ $\star$ $\star$ $q_3$)
\end{itemize}
\end{multicols}
\noindent where $q_1$, $q_2$, and $q_3$ are the words of the triple in their
linear order, and $\star$ is a single intervening word of any kind.
% Siblings
% are summed as per the first-order features described in
% Section~\ref{sect:first-order-surface-ngrams}.

We encode the \POS tags of the parent and children (or just the children for
sibling features), along with the bucketed count, directionality, and the
binned distance between the two children. We also extract paraphrase-style
features for siblings in the same way as for first-order \ngrams, and
cumulative variants up to the maximum bucket size.

\section{\label{sect:setup}Experimental Setup}

As with \citet{bansal:11a} and \citet{pitler:12b}, we convert the Penn
Treebank to dependencies using \pennconverter\footnote{\tiny
\url{http://nlp.cs.lth.se/software/treebank_converter/}} \citep{johansson:07a}
(henceforth \LTH) and generate \POS tags with \MXPOST \citep{ratnaparkhi:96a}. We used sections 02-21 of the \WSJ for training, 22 for
development, and 23 for final testing. The test sections of the answers,
newsgroups, and reviews sections of the English Web Treebank as per the \sancl
2012 Shared Task \citep{petrov:12b} were converted to \LTH and used for
out-of-domain evaluation.
We used \mstparser \citep{mcdonald:06a}, trained with the parameters
\emph{order:2}, \emph{training-k:5}, \emph{iters:10}, and
\emph{loss-type:nopunc}. We omit labeled attachment scores in this paper for
brevity, but they are consistent with the reported \UAS scores.

\section{\label{sect:results}Results}

\begin{table}[t]
  \centering
  \begin{small}
  \begin{tabular}{l|cccc|cc}
    \hline
                     &       &       &        &            & \BOOKS        & \hspace{-1mm}\% $\Delta$ \\
    \LTH             & \MST  & \WEBT & \BOOKS & \SYNTACTIC & \SYNTACTIC    & \hspace{-1mm}\MST \\
    \hline\hline
    \WSJ 22          & 92.3  & 92.9  & 92.9   & 92.7       & \textbf{93.2} & \hspace{-1mm}+0.9 \\
    \WSJ 23          & 91.7  & 92.2  & 92.3   & 92.4       & \textbf{92.6} & \hspace{-1mm}+0.9 \\
    \hline\hline
    \EWT \ANSWERS    & 82.5  & 83.4  & 83.2   & 83.6       & \textbf{83.6} & \hspace{-1mm}+1.1 \\
    \EWT \NEWSGROUPS & 85.2  & 86.1  & 86.1   & 86.1       & \textbf{86.4} & \hspace{-1mm}+0.9 \\
    \EWT \REVIEWS    & 83.6  & 84.5  & 84.3   & 84.9       & \textbf{85.0} & \hspace{-1mm}+1.3\\
    \hline
    \EWT \AVG        & 83.8  & 84.6  & 84.5   & 84.8       & \textbf{85.0} & \hspace{-1mm}+1.2 \\
    \hline
  \end{tabular}
  \end{small}
  \caption{\label{table:lth-results} \LTH \UAS (\mstparser) on the \WSJ dev
    and test set, and English Web
  Treebank (\EWT) answers (\ANSWERS), newsgroups (\NEWSGROUPS), and reviews
  (\REVIEWS) test set for the baseline (\BASE),
  Web1T (\WEBT), Google Books (\BOOKS), Syntactic (\SYNTACTIC), and combined
  (\BOOKS + \SYNTACTIC) feature sets. All results are statistically significant
  improvements over the baseline.}
\end{table}

% \begin{table}[t]
%   \centering
%   \begin{tabular}{l|rrrr|r}
%     \hline
%                      &       &       &        &            & \BOOKS        \\
%     Stanford         & \BASE & \WEBT & \BOOKS & \SYNTACTIC & \SYNTACTIC    \\
%     \hline\hline
%     \WSJ 22          & 91.2  & 91.6  & 91.9   & 91.6       & \textbf{92.0} \\
%     \WSJ 23          & 90.7  & 91.5  & 91.5   & 91.6       & \textbf{92.0} \\
%     \hline\hline
%     \EWT \ANSWERS    & 80.3  & 81.2  & 81.2   & 81.4       & \textbf{81.7} \\
%     \EWT \NEWSGROUPS & 84.1  & 85.2  & 85.5   & 85.1       & \textbf{85.8} \\
%     \EWT \REVIEWS    & 81.2  & 82.0  & 82.0   & 82.5       & \textbf{82.9} \\
%     \hline
%     \EWT \AVG        & 81.9  & 82.8  & 82.9   & 83.0       & \textbf{83.5} \\
%     \hline
%   \end{tabular}
%   \caption{\label{table:stanford-results} Stanford \UAS on the \WSJ and
%   English Web Treebank (\EWT) answers (\ANSWERS), newsgroups (\NEWSGROUPS), and
%   reviews (\REVIEWS) test corpora. All results are statistically significantly
%   improvements over the baseline.}
% \end{table}

% NN result: google parser not having NP bracketing

Table~\ref{table:lth-results} summarizes our results over the \WSJ development
and test datasets, and the \sancl 2012 test datasets. All of our features
perform very similarly to one another: each feature set in isolation provides
a roughly 0.5\% \UAS improvement over the baseline parser on the \WSJ
development and test sections. On the out-of-domain web treebank, surface and
syntactic features each improve over the baseline by an average of roughly 0.8
-- 1.0\% on the test sets. All of our results are also statistically
significant improvements over the baseline.

While our syntactic \ngram counts are computed over Stanford dependencies and
almost certainly include substantial parser and \OCR errors, they still
provide a significant performance improvement for \LTH parsing. Additionally,
the \syntacticngrams dataset is drawn from a wide variety of genres, but helps
with both newswire and web text parsing.
% \SYNTACTIC column in Tables~\ref{table:lth-results}
% and~\ref{table:stanford-results} show these features produced roughly
% equivalent absolute \UAS improvements in- and out-of-domain on both schemes.
% We regard this as further evidence for the \citet{sogaard:13a} hypothesis
% that differences between dependency schemes are smoothed by parser bias.

% \subsection{\LTH Dependencies}

The best results on \LTH dependencies used second-order sibling features in
addition to the first-order features for both surface and syntactic \ngrams. A
combined system of Google Books surface \ngram features and syntactic \ngram
features (which performed individually best on the development set) produces
absolute \UAS improvements of 0.8\% over the baseline on the \WSJ test set,
and 1.4\% over the baseline averaged across the three web treebank testing
domains. These results are significantly higher than any feature set in
isolation, showing that surface and syntactic \ngram features are
complementary and individually address different types of errors being made by
the parser.

% \subsection{Stanford Dependencies}

% On Stanford dependencies, we found that our surface \ngram system benefited
% from sibling features, but second-order syntactic \ngram features did not
% provide a statistically significant improvement. Thus, our best results in
% Table~\ref{table:stanford-results} include second-order siblings features for
% surface \ngram systems, but only first-order features for the syntactic \ngram
% system. Our combined system uses surface \ngrams from Google Books as these
% performed best on the Stanford development set; this system produces a 1.3\%
% absolute improvement over the \WSJ test set and 1.6\% averaged over the web
% treebank test sets. The lower results on the Stanford scheme compared to \LTH
% are partially explained by the smaller \ontonotes training corpus.

\section{\label{sect:analysis}Analysis}

% without being able to evaluate the Google parser accuracy, it is hard to
% know whether the longer deps are worse for syntax because the parser is bad
% at them

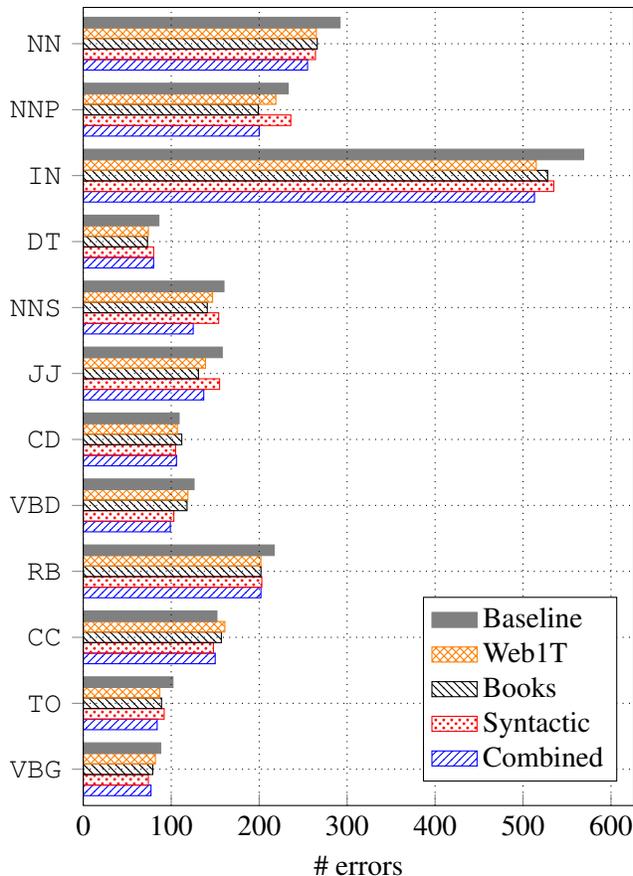
\begin{figure}[t]
    \begin{tikzpicture}
    \tikzstyle{delta}=[above,minimum height=0,text height=0.5ex, text depth=0ex,inner sep=1, outer sep=1,font=\scriptsize]
    \begin{axis}[
        xbar=0pt,% space of 0pt between adjacent bars
        xmin=0,
        xtick pos=left,
        ytick pos=left,
        y tick label style={font=\ttfamily},
        bar width=4,
        width=8.9cm,
        height=12.2cm,
        grid style={dotted,black},
        grid=both,
        minor y tick num=0,
        ytick=data,
        legend style={at={(0.5,-0.1)},
          anchor=north,legend columns=1},
        xlabel={\# errors},
        symbolic y coords={ VBG,TO,CC,RB,VBD,CD,JJ,NNS,DT,IN,NNP,NN, },
        % nodes near coords, nodes near coords align={horizontal},
        enlarge y limits=0.05,
        area style,
        reverse legend,
        legend cell align=left,
        legend pos=south east
      ]
      \addplot [draw=blue,pattern color=blue, pattern=north east lines] coordinates {(77,VBG) (84,TO) (150,CC) (202,RB) (99,VBD) (106,CD) (137,JJ) (125,NNS) (80,DT) (513,IN) (200,NNP) (255,NN)  };
      \addplot [draw=red,pattern color=red,pattern=crosshatch dots] coordinates { (74,VBG) (92,TO) (148,CC) (203,RB) (103,VBD) (105,CD) (155,JJ) (154,NNS) (80,DT) (535,IN) (236,NNP) (264,NN)  };
      \addplot [draw=black, pattern=north west lines] coordinates {(79,VBG) (89,TO) (157,CC) (202,RB) (118,VBD) (112,CD) (131,JJ) (141,NNS) (73,DT) (528,IN) (199,NNP) (266,NN)  };
      \addplot [draw=orange, pattern color=orange, pattern=crosshatch] coordinates {(82,VBG) (87,TO) (161,CC) (202,RB) (119,VBD) (107,CD) (139,JJ) (147,NNS) (74,DT) (515,IN) (219,NNP) (265,NN)  };
      \addplot [draw=gray,fill=gray] coordinates { (88,VBG) (102,TO) (152,CC) (217,RB) (126,VBD) (109,CD) (158,JJ) (160,NNS) (86,DT) (569,IN) (233,NNP) (292,NN)  };
      \legend{Combined,Syntactic,Books,Web1T,Baseline}
\end{axis}
\end{tikzpicture}
  \caption{\label{figure:errors}Total \LTH attachment errors by
  gold argument \POS tag, sorted by the total tag frequency.}
\end{figure}

Figure~\ref{figure:errors} gives an error breakdown by high-frequency gold
argument \POS tag on \LTH dependencies for the baseline, Web1T surface
\ngrams, syntactic \ngrams, and combined systems reported in
Table~\ref{table:lth-results}. For almost every \POS tag, the combined system
outperforms the baseline and makes equal or fewer errors than either the
surface or syntactic \ngram features in isolation. Syntactic \ngrams are worse
relative to surface \ngrams on noun, adjectival, and prepositional parts of
speech -- constructions which are known to be difficult to parse. Without
\NP-bracketed training data or the extra features that we have discussed as
helping resolve these issues, it is unsurprising that syntactic \ngram
features using the counts from the \citet{goldberg:13a} parser are less
effective. In comparison, surface \ngrams are worse on conjunctive and verbal
parts of speech, suggesting that the localized nature of these features is
less useful for the idiosyncrasies of coordination representations and
longer-range subject/object relationships.

Whilst Web1T and Google Books features perform similarly overall, Books
\ngrams are more effective for noun structures, and Web1T \ngrams are slightly
better in predicting \PP attachment sites.

% verbs: root, longer range
% conjunction: weird structure

\begin{table}[t]
\begin{center}
  \begin{tabular}{l|r|r|r|r}
    \hline
    Tag            & Freq & \BASE        & \COMBINED     & \%   \\
    \hline\hline
    \texttt{NN}    & 5725 & 5433         & \textbf{5470} & 12.0 \\
    \texttt{NNP}   & 4043 & 3810         & \textbf{3843} & 10.7 \\
    \texttt{IN}    & 4026 & 3457         & \textbf{3513} & 18.2 \\
    \texttt{DT}    & 3511 & 3425         & \textbf{3431} & 2.0  \\
    \texttt{NNS}   & 2504 & 2344         & \textbf{2379} & 11.4 \\
    \texttt{JJ}    & 2472 & 2314         & \textbf{2335} & 6.8  \\
    \texttt{CD}    & 1845 & 1736         & \textbf{1739} & 1.0  \\
    \texttt{VBD}   & 1705 & 1579         & \textbf{1606} & 8.8  \\
    \texttt{RB}    & 1308 & 1091         & \textbf{1106} & 4.9  \\
    \texttt{CC}    & 1000 & 848          & \textbf{850}  & 0.7  \\
    \texttt{VB}    & 983  & 941          & \textbf{947}  & 2.0  \\
    \texttt{TO}    & 868  & 766          & \textbf{784}  & 5.8  \\
    \texttt{VBN}   & 850  & 783          & \textbf{792}  & 2.9  \\
    \texttt{VBZ}   & 705  & 636          & \textbf{638}  & 0.7  \\
    \texttt{PRP}   & 612  & 604          & \textbf{606}  & 0.7  \\
    \texttt{VBG}   & 588  & 500          & \textbf{511}  & 3.6  \\
    \texttt{POS}   & 428  & 422          & 422           & 0.0  \\
    \texttt{\$}    & 352  & \textbf{345} & 343           & -0.7 \\
    \texttt{MD}    & 344  & 307          & \textbf{313}  & 2.0  \\
    \texttt{VBP}   & 341  & 298          & \textbf{305}  & 2.3  \\
    \texttt{PRP\$} & 288  & \textbf{281} & 280           & -0.3 \\
    Other          & 1010 & 868          & \textbf{883}  & 4.9  \\
    \hline
  \end{tabular}
  \caption{\label{table:combined-pos-tag-errors}Correct attachments by gold
  argument \POS tag and the percentage of the overall error reduction over
  \WSJ section 22 for the baseline and combined systems in
  Table~\ref{table:lth-results}.}
\end{center}
\end{table}

Table~\ref{table:combined-pos-tag-errors} lists a complete breakdown of
correct attachments corrected by the combined system. The most substantial
gains come in nominal and prepositional phrases -- known weaknesses for
parsers, and the categories where syntactic \ngram features alone fare worst.
However, the system finds less improvement in coordinators, determiners, and
cardinal numbers, all of which are also components of noun phrases. This shows
the difficulty of correctly identifying a head noun in a nominal to attach
modifiers to, and the general difficulty of representing and parsing
coordination.

\begin{table}[t]
\begin{center}
  \begin{tabular}{l|r|r}
    \hline
    Corpus        & Not Present & \%   \\
    \hline\hline
    Google Books  & 1,714,631   & 32.5 \\
    Web1T         & 1,425,347   & 27.0 \\
    \hline
    Intersection  & 1,301,090   & 24.7 \\
    \hline
  \end{tabular}
  \caption{\label{table:web1t-books-missing-counts}Surface \ngram
  queries from the \WSJ and English Web Treebank that do not receive features
  from Web1T and Google Books.}
\end{center}
\end{table}

% % complementary coverage

Web1T contains approximately double the total number of \ngrams as Google
Books. Table~\ref{table:web1t-books-missing-counts} shows that 27\% and 32.5\%
of the \ngram queries from the \WSJ sections 2-23 and the entire English Web
Treebank do not receive features from Web1T and Google Books respectively. The
intersection of these queries is 24.7\% of the total, showing that the two
corpora have small but substantive differences in word distributions; this may
partially explain why our combined feature experiments work so well. However,
the similar performance of surface \ngram features extracted from these sources
suggests Web1T contains substantial noise.

We had expected our syntactic \ngram features to perform better than they did
since they address many of the shortcomings of using surface \ngrams.
Syntactic features are sensitive to the quality of the parser used to produce
them, but in this case the parser is difficult to assess as the source corpus
is enormous and extracted using \OCR from scanned books. Even if the parser is
state of the art, it is being used to parse diverse texts spanning multiple
genres across a wide time period, compounded by potential scanning and
digitization errors. Additionally, a post-hoc analysis of the types of errors
present in the corpus is impossible due to the exclusion of the full parse
trees, though \citet{goldberg:13a} note that this data would almost certainly
be computationally prohibitive to process. Despite this, our work has shown
that counts from this corpus provide useful features for parsing. Futhermore,
these features stack with surface \ngram features, providing substantial
overall performance improvements.

\subsection{Future Work}

A combination of features from all of the sources used in this work would be
interesting avenues for further investigation, especially since these features
seem strongly complementary. We could also explore more of the \POS and
head-modifier annotations available in the Google Books Ngram corpus to
develop features which are a middle ground between surface and syntactic
\ngram features.

The Google Books and \syntacticngrams corpora both provide frequencies by
date, and it would be interesting to explore how well features extracted from
different date ranges would perform -- particularly on text from roughly the
same eras. Resampling Web1T to reduce it to a comparable corpus that is the
same size as Google Books would also provide better insights on how many
\ngrams are noise.

% Stanford vs. LTH: SUBJ/nsubj, VC/aux
% too many numbers
% just the final numbers on test set
% LTH and Stanford in the same float
% captions too big

% \begin{table}
%   \begin{tabular}{l|r|r|r|r}
% Length & Freq & Web1T & Syn & Comb\\\hline\hline
% 1  & 18766 & 18031 & 18043 & 18070 \\
% 2  & 6880  & 6452  & 6446  & 6459  \\
% 3  & 3566  & 3272  & 3244  & 3281  \\
% 4  & 1847  & 1602  & 1593  & 1613  \\
% 5  & 1037  & 855   & 851   & 860   \\
% 6+ & 3412  & 2781  & 2781  & 2813
%   \end{tabular}
%   \caption{\label{table:length-comparison}A comparison of correct dependencies
%   by length for surface Web1T features, syntactic \ngram features, and the
%   combination of the two.}
% \end{table}

\section{\label{sect:background}Related Work}

% The use of automatically parsed data for improving parser accuracy has been
% explored in several different ways. \citet{sarkar:01a} and
% \citet{steedman:03a} have applied co-training techniques to constituency
% parsing, finding that it  is best suited to situations where there is only a
% small amount of labeled data available. \citet{sagae:07a} have also found
% co-training to be useful for domain adaptation, using parsed out-of-domain
% data to augment annotated training data.

% \citet{mcclosky:06a} self-trained the Charniak parser and reranker on its own
% output on in-domain newswire text, producing substantial performance
% improvements on the \WSJ. The interaction between the reranker and parser
% appears to be crucial in producing sufficiently differentiated parses for
% self-training to succeed.

% In co-training, two systems are iteratively trained on a small amount of
% annotated data augmented with the most confident predictions of the other
% system, allowing each to benefit from the insights of the other.

% We also make use of the extra \POS tag information provided in the corpus for our
% features.

% "As our approach extends upon Bansal and Klein (2011), we describe the details
% of this work below as we describe our extensions."

Surface \ngram counts from large web corpora have been used to address \NP and
\PP attachment errors \citep{volk:01a,nakov:05a} Aside from
\citet{bansal:11a}, other feature-based approaches to improving dependency
parsing include \citet{pitler:12b}, who exploits Brown clusters and point-wise
mutual information of surface \ngram counts to specifically address \PP and
coordination errors. \citet{chen:13a} describe a novel way of generating
meta-features that work to emphasise important feature types used by the
parser.

\citet{chen:09a} generate subtree-based features that are similar to ours.
However, they use the in-domain \BLLIP newswire corpus to generate their
subtree counts, whereas the \syntacticngrams corpus is out-of-domain and an
order of magnitude larger. They also use the same underlying parser to
generate the \BLLIP subtree counts and as the final test-time parser, while
\syntacticngrams is parsed with a simpler, shift-reduce parser compared to the
graph-based \mstparser used during test time. They also evaluate only on
newswire text, whilst our work systematically explores various
configurations of surface and syntactic \ngram features in- and out-of-domain.

% This work is the first to introduce count-based subtree features from
% a corpus of the size of Google \syntacticngrams.

\section{\label{sect:conclusion}Conclusion}

We developed features for dependency parsing using subtree counts from 345
billion parsed words of scanned English books. We extended existing work on
surface \ngrams from first to second-order, and investigated the utility of
web text and scanned books as sources of surface \ngrams.

Our individual feature sets all perform similarly, providing significant
improvements in parsing accuracy of about 0.5\% on newswire and up to 1.0\%
averaged across the web treebank domains. They are also complementary, with
our best system combining surface and syntactic \ngram features to achieve up
to 1.3\% \UAS improvements on newswire and 1.6\% on web text. We hope that our
work will encourage further efforts to unify different sources of unlabeled
and automatically parsed data for dependency parsing, addressing the relative
strengths and weaknesses of each source.

\bibliographystyle{acl}
{\small
\bibliography{naacl15ngrams}
}

\end{document}